\PassOptionsToPackage{table}{xcolor}
\documentclass[onecolumn]{fairmeta}

\usepackage{hyperref}

\usepackage{amsmath}
\usepackage{amssymb}
\usepackage{mathtools}
\usepackage{bm}
\usepackage{microtype}
\usepackage{graphicx}
\usepackage{subcaption}
\usepackage{booktabs}
\usepackage{epigraph}
\usepackage{xspace}
\usepackage{xstring}
\usepackage{array}
\usepackage{algorithm}
\usepackage{algorithmic}
\usepackage{amsmath}

\usepackage{wrapfig}
\usepackage{array}
\usepackage{nicefrac}
\usepackage{amsfonts}
\usepackage{url}

\usepackage{wrapfig} 
\usepackage{lipsum}  

\usepackage{tcolorbox}
\tcbuselibrary{breakable, skins}

\newtcolorbox{rubricbox}{
  enhanced, breakable,
  colback=blue!4,           
  colframe=blue!55!black,   
  coltitle=white,
  fonttitle=\bfseries,
  title={AR-Eval Rubric (judge instruction)},
  boxrule=0.5pt, arc=2pt,
  left=6pt, right=6pt, top=6pt, bottom=6pt,
}
\crefname{table}{Table}{Tables}
\Crefname{table}{Table}{Tables}

\crefname{section}{Section}{Sections}
\Crefname{section}{Section}{Sections}
\crefname{figure}{Figure}{Figures}
\Crefname{figure}{Figure}{Figures}

\usepackage{nicematrix}
\newcommand{\pmSep}{%
  \ifnum\theiRow>2\relax 
    $\,\pm\,$%
  \fi
}
\usepackage{multirow}

\usepackage{tcolorbox}
\tcbuselibrary{skins}
\tcbuselibrary{breakable}
\usepackage{float}
\usepackage{siunitx}
\newcommand{\circa}{{\raise.17ex\hbox{$\scriptstyle\sim$}}}
\usepackage{graphicx} 
\usepackage{booktabs}
\usepackage{pifont}

\usepackage[textsize=tiny]{todonotes}

\usepackage{tikz}
\usetikzlibrary{patterns}

\newtcolorbox{questionbox}[1]{
  colback=metabg,
  colframe=metafg,
  fonttitle=\bfseries,
  title=#1
}

\definecolor{icmlblue}{RGB}{0,76,153} 
\definecolor{icmlgray}{gray}{0.15}
\definecolor{icmllight}{gray}{0.96}

\tcbset{
  icmlbox/.style={
    enhanced,
    breakable,
    boxrule=0.4pt,
    colframe=icmlblue!65!black,
    colback=icmllight,
    coltitle=black,
    fonttitle=\bfseries,
    left=1.0mm,right=1.0mm,top=0.8mm,bottom=0.8mm,
    boxsep=1.2mm,
    arc=1.2mm,
    outer arc=1.2mm,
    titlerule=0pt,
    toptitle=0.4mm,
    bottomtitle=0.2mm,
    attach title to upper,
    title={#1},
  },
  icmlbox/.default={}
}
\newtcolorbox{callout}[1][]{%
  enhanced, breakable, frame hidden,
  left=2mm,right=2mm,top=2mm,bottom=2mm,
  boxrule=1pt,
  colback=metabg,
  arc=10pt,
  before skip=15pt,
  grow to left by=3pt,
  grow to right by=3pt,
  #1
}

\title{ 
FastJEV: Understanding Redundancy for Compact JEV Inference
}

\author{Jie Ma}
\author{Jie Gao}
\author{Yihang Liu}
\author{Zhike Qiu}
\author{Junle Li}
\author{Chongyi Zhuang}
\author{Jiayi Ji}
\author{Xiaoshuai Sun}

\affiliation{Xiamen University}

\abstract{
JEV models make multimodal decisions by directly scoring candidates. Although the common context is encoded once, candidate evaluation can still repeat matching token histories, duplicate inference states, and execute the full backbone. 
In this paper, we study these sources of redundancy and present FastJEV for compact candidate evaluation. We jointly organize history reuse and state storage, since sharing computation requires preserving states for later branches. We first introduce shared context anchoring to reuse recurrent initial states and omit unused final recurrent caches. We extend this reuse through candidate prefix sharing, retaining the intermediate states needed by subsequent branches. To further reduce the depth of these paths, we apply decision guided pruning based on relative score changes measured on a small unlabeled set. Our method retains full context encoding and all candidates without additional training.
We evaluate FastJEV across three OmniJev model sizes on five public benchmarks and reconstructed LIBERO-10 offline questions. At the selected pruning budgets, the complete method reduces candidate depth by 43.75\% to 45.83\%, while retaining 93.66\% to 97.52\% of the original task scores on average across the six evaluation sets.
Through controlled experiments, we show how candidate overlap and branching structure affect the execution cost of history reuse. In our implementation, candidate prefix sharing can reduce repeated computation while increasing latency. These findings motivate designing sharing granularity and execution schedules together for efficient JEV inference.
}

\date{\today}
\correspondence{\email{jiema100@stu.xmu.edu.cn},  \email{xssun@xmu.edu.cn}
}
\metadata[Code]{\url{https://github.com/JieMaMagic/FastJEV}}

\begin{document}

\maketitle

\section{Introduction}
\label{sec:introduction}

Multimodal models connect visual perception with language understanding, supporting tasks such as recognition, question answering, and instruction following~\citep{clip2021,llava2023,blip22023}.
Their capabilities also extend to embodied settings, where visual observations and language instructions guide action selection~\citep{rt22023,geminirobotics2025}.
Across these settings, a central role of multimodal models is to turn observations into task decisions.
JEV makes this role explicit by directly scoring candidate choices without autoregressive answer generation~\citep{omnijev2026}.
This approach places decision computation at the center of inference.
Understanding how that computation can be organized compactly is an important step toward efficient multimodal decision models.

Research on efficient inference has explored both computation reuse and selective reduction.
One line of work shares common prefixes and cached states to reduce repeated computation and duplicate storage~\citep{pagedattention2023,sglang2024}.
Another reduces computation according to estimated redundancy, either by pruning layers~\citep{shortgpt2025,sleb2024} or by shortening visual token sequences~\citep{fastv2024}.
More selective strategies use output sensitivity to bypass visual attention operations while preserving feed-forward updates~\citep{vskip2026}.
Together, these directions motivate studying how computation can be shared and reduced within multimodal decision models.

In direct candidate scoring, a common multimodal context can be encoded once and reused across alternatives~\citep{omnijev2026}.
Candidate paths can still repeat matching histories, duplicate inference states, and execute the full backbone.
These costs are closely related.
Reusing a history removes repeated computation but requires retaining states for its later branches.
Reducing depth changes the computation along both shared paths and their continuations, and its effect must be assessed through the final candidate scores.
\textit{How can we organize candidate histories, inference states, and execution depth into a compact structure while preserving decision quality?}

To address this question, we present FastJEV, which connects shared context anchoring, candidate prefix sharing, and decision guided pruning.
We first use the context as a common anchor, sharing its recurrent initial states across candidate paths and omitting unused final recurrent caches.
Candidate prefix sharing extends this reuse through a trie of matching token histories.
Each shared history is evaluated once, and states needed by later branches are retained.
We then select layers for pruning by measuring how skipping each layer changes relative candidate scores.
A small unlabeled calibration set provides a fixed layer ranking for each model.
For each pruning budget, we apply the selected mask to every candidate path and reuse it across tasks.
FastJEV retains full context encoding and all answer candidates without additional training.

We evaluate FastJEV using the 0.8B, 2B, and 4B models of OmniJev on five public benchmarks~\citep{ai2d2016,scienceqa2022,pope2023,blink2024,geminirobotics2025} and reconstructed LIBERO-10 offline questions~\citep{libero2023}.
At the selected pruning budgets, the complete method reduces candidate depth by 43.75\% to 45.83\%.
These configurations retain 93.66\% to 97.52\% of the original task scores on average across the six evaluation sets.
Controlled experiments show how history sharing reduces repeated candidate computation and how state management reduces live state storage.
They further reveal that execution cost depends on candidate overlap and branching structure.
With independent batched candidate evaluation, shared context anchoring and pruning achieve speedups of $1.053\times$ to $1.258\times$ at the default six-layer pruning budget on one NVIDIA L20 GPU.
These findings provide a structural view of JEV inference and clarify when compact candidate evaluation improves execution efficiency.

Our contributions are summarized as follows.
\begin{itemize}
  \item We provide a structural analysis of JEV candidate evaluation that connects repeated token histories, inference state storage, and execution depth.

  \item We present FastJEV, which combines shared context anchoring, candidate prefix sharing, and decision guided pruning for compact candidate evaluation. The method retains full context encoding and all candidates without additional training.

  \item We evaluate FastJEV across three model sizes and six evaluation sets.
    At the selected pruning budgets, the complete method reduces candidate depth by 43.75\% to 45.83\%, with mean task score retention of 93.66\% to 97.52\%.
    Controlled analyses examine how candidate overlap and branching structure affect execution cost.
\end{itemize}
\section{Related Work}
\label{sec:related_work}

\subsection{Efficient Multimodal Inference}

Multimodal inference can produce decisions through representation matching~\citep{clip2021} or autoregressive generation~\citep{llava2023,blip22023}.
Direct candidate scoring provides another formulation, in which a shared context supports the comparison of explicit alternatives~\citep{omnijev2026}.
These formulations expose different sources of inference cost.
For generative vision-language models, selective execution often targets the visual inputs.
Attention importance can guide visual token pruning~\citep{fastv2024}, while output sensitivity can guide selective skipping of visual attention~\citep{vskip2026}.
FastJEV studies the computation required to compare candidates after a common context has been encoded.
It retains full context processing and all candidates, and organizes the remaining computation through history sharing, state reuse, and depth reduction.

\subsection{Prefix and State Sharing}

Prefix reuse reduces repeated computation when multiple requests or branches share an input history.
Block-based KV cache management supports shared storage across requests and decoding branches~\citep{pagedattention2023}.
Tree-based prefix indexing extends cache reuse to structured language model programs~\citep{sglang2024}.
In hybrid backbones, recurrent layers introduce state representations beyond attention KV caches~\citep{gateddeltanet2025}.
Recent work compresses the recurrent state information needed for speculative verification and reconstructs the accepted state~\citep{treewy2026}.
FastJEV connects history reuse and state management within a candidate scoring request.
The common context supplies a shared recurrent anchor, and a candidate trie extends reuse to matching suffix histories.
State retention follows the tree dependencies, preserving continuation states and omitting unused final recurrent caches.
This organization links the computation shared by candidates to the states needed for their subsequent evaluation.

\subsection{Structured Pruning}

Structured pruning reduces computation by removing coupled parameter groups or complete layers.
One approach uses gradient information to prune coupled structures, followed by parameter-efficient recovery~\citep{llmpruner2023}.
Layer-level approaches estimate redundancy from changes in hidden representations~\citep{shortgpt2025} or changes in predictive loss after block removal~\citep{sleb2024}. Output sensitivity also guides visual attention skipping through divergence between vocabulary distributions, while feed-forward updates are retained~\citep{vskip2026}. FastJEV follows this sensitivity-based perspective, with relative candidate scores as the decision signal.
We rank layers using a small unlabeled calibration set and bypass selected complete blocks only during candidate evaluation.
The same selected layers apply to shared histories and their continuations, connecting depth reduction with prefix and state reuse.
Full context encoding is retained, and the method requires no additional training.

\section{Method}
\label{sec:method}

\begin{figure*}[!t]
    \centering
    \includegraphics[width=\textwidth]{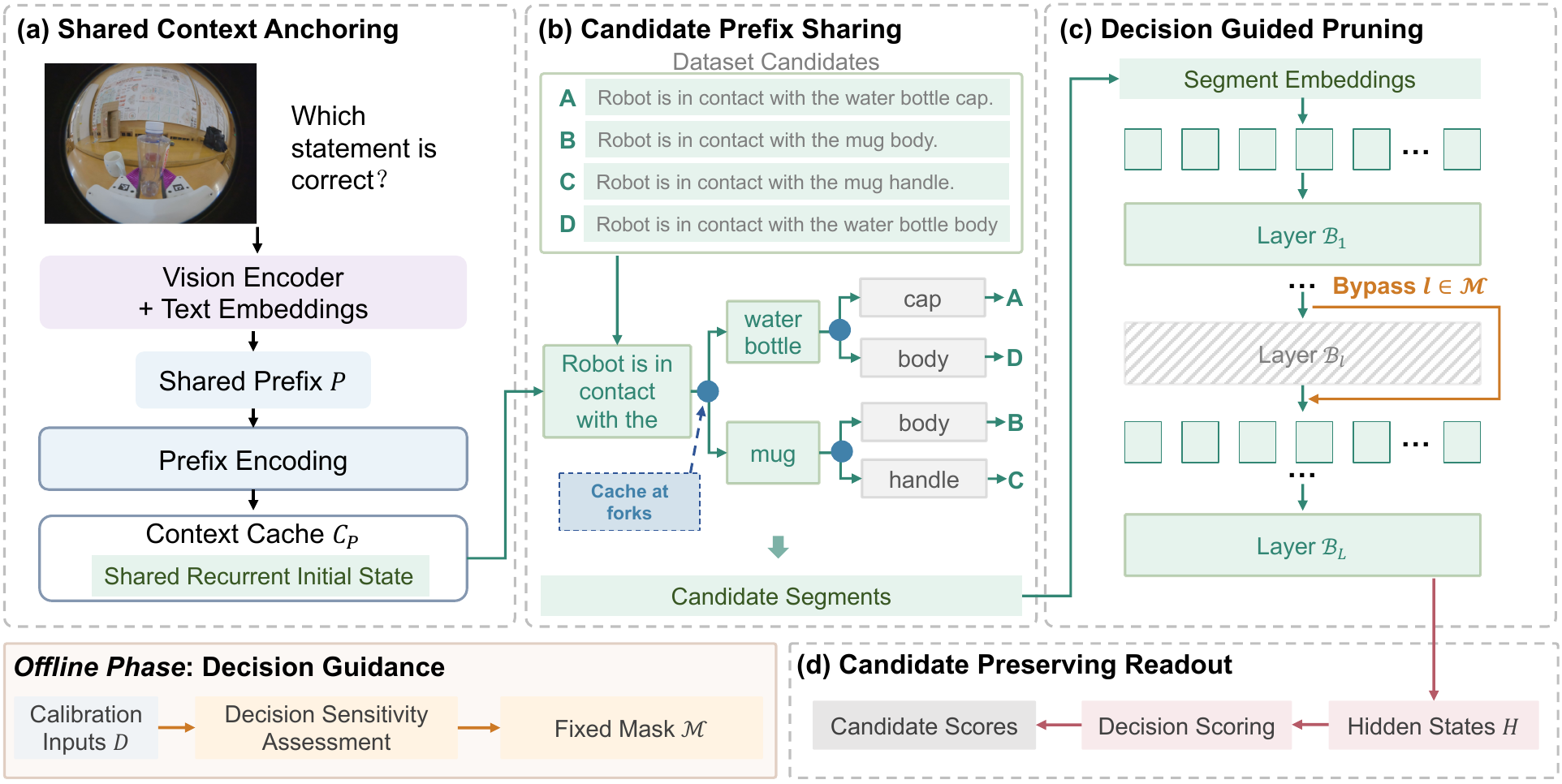}
    \caption{
        \textbf{Overview of FastJEV}, illustrated with an ERQA
        example~\citep{geminirobotics2025} on
        OmniJev~\citep{omnijev2026}.
        (a) Shared context anchoring reuses the encoded context
        and shares recurrent initial state storage.
        (b) Candidate prefix sharing merges matching token
        histories and retains continuation caches at forks.
        (c) Decision guided pruning bypasses selected candidate
        layers using a fixed mask determined through offline
        sensitivity assessment.
        (d) The original scoring procedure produces candidate
        logits in their original order and an abstention logit.
        Full context encoding and all candidate tokens are retained.
        Phrase boxes group consecutive tokens, and layer positions
        are schematic.
    }
    \label{fig:method}
\end{figure*}

\paragraph{Overview.}
FastJEV organizes candidate evaluation into a compact structure, as illustrated in \cref{fig:method}. We connect the reuse of states and matching histories with a reduction in candidate execution depth. Shared context anchoring (\cref{sec:context_anchoring}) retains full context encoding and shares recurrent initial state storage across candidate paths. However, starting from the same context does not eliminate repeated computation over matching candidate prefixes. Candidate prefix sharing (\cref{sec:candidate_prefix_sharing}) therefore extends reuse beyond the initial context by evaluating matching token histories only once. This sharing removes duplicate prefix computation, but each remaining segment still traverses the full backbone. Decision guided pruning (\cref{sec:decision_pruning}) then reduces candidate depth by bypassing layers selected according to their effect on final scores.

\subsection{Preliminaries on JEV}
\label{sec:jev_background}

For visual tasks, OmniJev combines a visual encoder, a language backbone, and a decision head~\citep{omnijev2026}.
The visual encoder maps images to features that the language backbone processes with the question and answer candidates.
The decision head then assigns a score to each candidate.
We use OmniJev, whose Qwen3.5 backbone combines full attention and Gated DeltaNet layers~\citep{gateddeltanet2025}.

Given images $I$, a question $q$, and candidates $\mathcal A=\{a_i\}_{i=1}^{N}$, OmniJev forms a shared prefix $P$ and candidate suffixes $R_i$.
The prefix contains the shared visual and text context.
Each suffix contains the tokens needed to score one candidate, including its boundary markers.
Each suffix retains the token position used to read out the question representation.
The full hybrid backbone encodes $P$ once and produces a prefix cache $\mathcal C_P$.
For candidate $i$, we compute the hidden outputs from the prefix cache as
\begin{equation}
H_i=F_{\theta}(R_i;\mathcal C_P),
\label{eq:jev_branch}
\end{equation}
where $F_{\theta}$ denotes the language backbone with pretrained weights $\theta$, and $H_i$ contains the hidden outputs of branch $i$.
The original readout $g_{\theta}$ combines question and candidate representations with language model features to produce
\begin{equation}
\mathbf z=g_{\theta}(H_1,\ldots,H_N).
\label{eq:jev_scoring}
\end{equation}
For multiple choice, $\mathbf z\in\mathbb R^{N+1}$ contains the candidate logits and one abstention logit.
The latter is produced by the head without an additional backbone branch.

This structure separates shared context encoding from the work needed to score each candidate.
The hybrid implementation reuses prefix computation but copies its recurrent states when creating branches.
Matching token histories within candidate suffixes can also incur repeated computation.
Every branch also runs all $L$ language layers.
We use this separation to organize candidate histories, state storage, and execution depth around a fully encoded context.

\subsection{Shared Context Anchoring}
\label{sec:context_anchoring}

On the hybrid backbone, all candidates begin with the same encoded history.
We treat the complete prefix cache as a common anchor and extend its reuse from prefix computation to recurrent state storage.
The anchor is the cache already produced by the original full backbone.
It requires no learned parameters or additional context encoding.

Let $S_P^{\ell}$ be the recurrent state produced by the prefix at Gated DeltaNet layer $\ell$.
We set the initial state of candidate $i$ at layer $\ell$ as
\begin{equation}
S_{i,0}^{\ell}=S_P^{\ell}.
\label{eq:context_anchor}
\end{equation}
We implement this common initial condition through a single read-only buffer.
Each branch reads the buffer and independently updates its own working state.
Sharing is valid when the visual and token history, positions, and masks are identical.
The initial value and branch updates match the original recurrence in exact arithmetic.
Convolution states and attention KV caches retain separate storage for each branch.

The anchor also clarifies which states must persist during scoring.
The prefix cache is retained because all candidate batches depend on it.
Once a candidate path ends and no continuation depends on it, its final recurrent cache has no further use.
We therefore omit its allocation and writeback, while still executing every recurrent update needed for scoring.
This includes the last update of each executed recurrent layer.
Intermediate caches needed to continue a candidate path are retained.

Context anchoring removes repeated state storage but still processes candidate suffixes separately.
When these suffixes contain matching histories, their token computations offer a further opportunity for reuse.
We next extend sharing to these histories while retaining the state needed at each fork.

\subsection{Candidate Prefix Sharing}
\label{sec:candidate_prefix_sharing}

Candidates may share an initial token sequence beyond the common context $P$.
Evaluating this sequence separately repeats the same transformation at each language layer.
We organize candidate suffixes into a prefix tree, or trie, and evaluate each shared node once per executed layer.
The tree starts from the common context anchor and carries the encoded history forward as candidate paths diverge.

\paragraph{History Consistent Merging.}
Let $r_{i,t}$ be the token at position $t$ of candidate suffix $R_i$, including its boundary markers.
Let $\mathbf p_{i,t}$ contain its three original position IDs, and let $v_{i,t-1}$ denote the tree node for its preceding history.
All suffixes start from the same root, which represents the common context $P$.
We assign the token to a node using the key
\begin{equation}
\kappa_{i,t}=\bigl(v_{i,t-1},r_{i,t},\mathbf p_{i,t}\bigr).
\label{eq:trie_key}
\end{equation}
Two occurrences share a node only when their keys are identical. The parent node identifies the complete preceding path, so equal tokens after different histories remain separate. Position IDs are retained from the unmerged input rather than reassigned after merging. All suffixes use the same starting position in the branch layout, so tree depth determines these position IDs.

\paragraph{Anchored Tree Execution.}
We execute the tree as path segments that end at a fork or a candidate boundary. Each segment starts from the cache for its parent history. This cache contains the attention keys and values, convolution states, and recurrent states of the retained layers. Segments with the same parent can be batched and share its recurrent state as their initial anchor. Each segment then updates its own working state. Segments with descendants retain their final caches for continuation. Only leaf segments may omit final recurrent cache storage.

The inherited cache and the causal mask within each segment restrict each token to $P$, its ancestors, and itself.
For a fixed set of executed layers, merged occurrences retain the same visible history and positions as in separate candidate evaluation.
Their hidden outputs are therefore unchanged in exact arithmetic during inference.
Finite precision can introduce differences when batch shapes change.

\paragraph{Candidate Preserving Readout.}
The decision head requires a separate entry for every candidate.
We retain a mapping from each token occurrence in $R_i$ to its tree node.
This mapping recovers the question and candidate representations required by $g_{\theta}$.
For the language model features, we project each required predictor node once and normalize over the full vocabulary.
We gather the token log probabilities along each candidate path to construct its original language model features.
The gathered representations and features enter the unchanged decision head in the original candidate order.
The head produces the candidate logits and the abstention logit.
Candidates that share a complete tree path remain separate entries in the readout.

\subsection{Decision Guided Pruning}
\label{sec:decision_pruning}

Prefix sharing reduces repeated evaluation of matching token histories, but leaves the depth of each candidate path unchanged.
We further reduce this depth by selecting layers according to their effect on final scores.
The resulting mask applies to all candidate paths, while shared context encoding retains the complete backbone.

\paragraph{Decision Sensitivity.}
The decision head combines several features.
Changes in intermediate representations alone do not specify their effect on candidate scores.
We therefore assess each layer through the output of the complete scoring process in Eq.~\eqref{eq:jev_scoring}.

Let $\mathcal D$ be a fixed calibration set of inputs $x=(I,q,\mathcal A)$.
Layer selection does not use the answer labels of these inputs.
For each $x\in\mathcal D$, let $\mathbf z(x)$ be the original logits and $\mathbf z^{(-\ell)}(x)$ the logits obtained by skipping only layer $\ell$ in candidate branches.
The prefix still uses every layer in both runs.
Both vectors include the abstention logit and are measured before probability postprocessing.
We write their difference as $\Delta_{\ell}(x)=\mathbf z^{(-\ell)}(x)-\mathbf z(x)$.

A common shift of all logits leaves their differences and softmax probabilities unchanged.
We remove this shift using $\operatorname{ctr}(v)=v-\operatorname{mean}(v)\mathbf 1$, where $\mathbf 1$ is the all-ones vector.
For an $m$-dimensional vector, let $\operatorname{rms}(v)=\lVert v\rVert_2/\sqrt m$.
We measure layer sensitivity by
\begin{equation}
d_{\ell}(x)=
\frac{\operatorname{rms}\!\left(\operatorname{ctr}(\Delta_{\ell}(x))\right)}
{\max\!\left(\epsilon,
\operatorname{rms}\!\left(\operatorname{ctr}(\mathbf z(x))\right)\right)},
\label{eq:layer_sensitivity}
\end{equation}
where $\epsilon=10^{-8}$.
The denominator accounts for score scale differences across inputs.
This criterion evaluates changes in relative scores, including changes that may leave the highest scoring candidate unchanged.

\paragraph{Calibration Guided Selection.}
The sensitivity $d_{\ell}(x)$ measures the effect of skipping layer $\ell$ for one input.
To select a fixed pruning mask, we average this quantity over $\mathcal D$ and define the layer score as
\begin{equation}
J_{\ell}=\frac{1}{|\mathcal D|}\sum_{x\in\mathcal D}d_{\ell}(x),
\label{eq:layer_importance}
\end{equation}
where $|\mathcal D|$ is the number of calibration inputs.
A smaller $J_{\ell}$ indicates a smaller average perturbation of relative candidate scores on $\mathcal D$.
Given a pruning budget $k$, we form $\mathcal M$ from the $k$ layers with the smallest $J_{\ell}$, with ties resolved by layer index.
We score layers individually and select all $k$ layers from the resulting ranking.
We select $\mathcal M$ once for each model and reuse it for subsequent inputs.
This calibration requires forward passes only and leaves the model weights unchanged.

\paragraph{Context Preserving Execution.}
Once $\mathcal M$ is fixed, we apply it only to candidate branches.
Shared segments and their continuations use the same mask.
The prefix continues to execute all layers and supplies the context cache for every retained block.
Let $H_i^{\ell}$ denote the hidden outputs of branch $i$ after layer $\ell$, with $H_i^0$ denoting its input embeddings.
For $\ell=1,\ldots,L$, we compute these outputs as
\begin{equation}
H_i^{\ell}=\begin{cases}
H_i^{\ell-1}, & \ell\in\mathcal M, \\
\mathcal B_{\ell}(H_i^{\ell-1};\mathcal C_P^{\ell}),
& \ell\notin\mathcal M,
\end{cases}
\label{eq:candidate_pruning}
\end{equation}
where $\mathcal B_{\ell}$ is the original language block and $\mathcal C_P^{\ell}$ is its prefix cache.
A selected block is replaced by an identity mapping, skipping both its attention or recurrent module and its feed-forward network.

At inference, FastJEV encodes the common context with the full backbone and uses its cache as the anchor for candidate evaluation.
Matching candidate histories share computation through the trie.
Each segment executes the retained blocks and preserves the states needed by its descendants.
The node mapping then recovers the features of every candidate for the original readout.
Together, these steps connect history sharing, state reuse, and depth reduction within a single scoring request.
All answer candidates remain available, and all model weights are retained for full context encoding.

\begin{figure*}[t]
\centering
\includegraphics[width=0.75\textwidth]{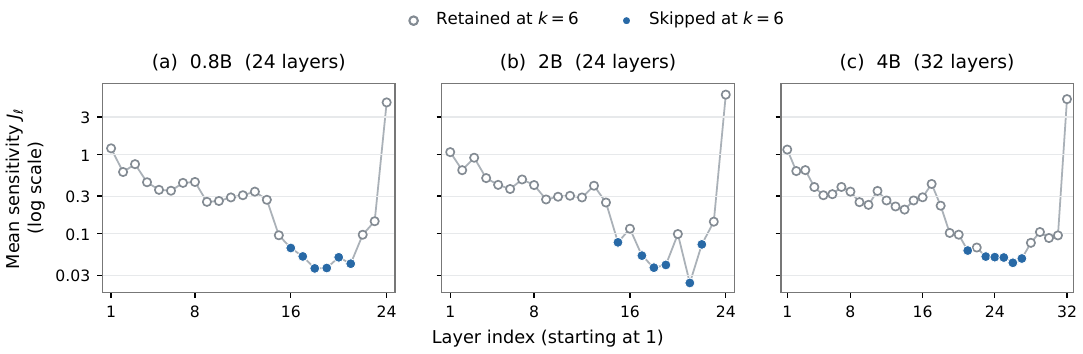}
\caption{Decision sensitivity across layers in OmniJev. Each point is the mean relative score perturbation $J_{\ell}$ over 32 multiple-choice calibration inputs when a single candidate block is skipped. The six lowest-scoring layers are highlighted for the default pruning budget. Layer indices start at one, and the vertical axis uses a logarithmic scale. Context encoding retains all layers.}
\label{fig:decision_sensitivity}
\end{figure*}

\paragraph{Empirical Justification.}
\cref{fig:decision_sensitivity} visualizes the calibration scores $J_{\ell}$ for three OmniJev model sizes.
Low sensitivity occurs in parts of the later backbone, while the final block has the highest sensitivity at every scale.
Thus, sensitivity does not decrease uniformly with depth.
This pattern motivates selecting layers by measured changes in relative candidate scores.

\section{Experiments}
\label{sec:experiments}

\subsection{Experimental Setup}
\label{sec:exp_setup}

\paragraph{Models and data.}
We use the released 0.8B, 2B, and 4B models of OmniJev v1.1~\citep{omnijev2026}.
The public benchmarks comprise AI2D test~\citep{ai2d2016}, ScienceQA-IMG test~\citep{scienceqa2022}, POPE~\citep{pope2023}, BLINK validation~\citep{blink2024}, and ERQA~\citep{geminirobotics2025}.
They contain 3,088, 2,017, 9,000, 1,901, and 400 questions, respectively.
We also construct an offline question set from LIBERO-10 demonstrations~\citep{libero2023}.
It contains 188 frames from 20 evaluation episodes across ten tasks, with eight question types per frame.
The 1,504 questions follow the public OmniJev question schema and available templates.
Their labels use simulator states and fixed trajectory rules.
The original question pool is unavailable, so LIBERO results measure offline question answering accuracy on this reconstructed set.

\paragraph{Configurations.}
T denotes candidate prefix sharing, S denotes shared context anchoring, and P denotes decision guided pruning.
Full FastJEV denotes T+S+P, and S+P is the ablation without T.
Original uses the released inference path, which already computes the common visual and text prefix once.
All configurations retain the original candidates and decision heads.
We compute one sensitivity ranking per model on 32 OK-VQA inputs~\citep{okvqa2019}, each with 64 fixed candidates drawn from the training answer vocabulary.
Layer ranking uses neither the calibration answer labels nor the target evaluation labels.
For each pruning budget, we select a fixed mask from this ranking and reuse it across tasks without further training.
The default budget skips six candidate blocks, reducing candidate depth by 25\% for 0.8B and 2B and by 18.75\% for 4B.
Larger budgets extend the same ranking.
The common prefix still executes every block, and all weights remain resident.

\paragraph{Evaluation protocol.}
We report accuracy on each complete evaluation split.
Public multiple-choice benchmarks select the highest-scoring provided option.
Binary questions use the original sigmoid head with a probability threshold of 0.5.
The LIBERO protocol includes abstention in the choice argmax and counts it as an error.
Mean retention is the unweighted mean of six accuracy ratios to the matched Original.
Ratios are computed before rounding and are not clipped at 100\%.

\paragraph{Cost measurement.}
Performance is measured with exclusive access to one NVIDIA L20 GPU using a BF16 backbone, SDPA, and the same PyTorch convolution backend.
Public benchmarks use 64 fixed questions for three timing rounds and 16 questions for memory measurement.
LIBERO uses 20 frames for three timing rounds and ten frames for memory measurement.
Each frame is evaluated through two requests covering its eight questions.
Latency includes image loading and preprocessing, context encoding, candidate execution, and scoring, with CUDA synchronization at the boundaries.
It excludes model loading, warmup, configuration installation, and result serialization.
We report the largest PyTorch peak allocated memory across the measured inputs, including model weights.
After warmup, configurations are timed in an interleaved order determined by a fixed random seed.

\subsection{Compact Candidate Evaluation across Tasks}
\label{sec:exp_main}

\begin{table*}[t]
\centering
\small
\setlength{\tabcolsep}{3pt}
\renewcommand{\arraystretch}{1.13}
\caption{Decision quality of OmniJev and FastJEV, grouped by model scale. All FastJEV rows use T+S+P. Pruning applies to candidate evaluation, while context encoding uses all layers. Dataset columns report accuracy (\%). Mean retention averages six accuracy ratios to the matched Original. Bold pruning and retention entries mark the largest tested budget with mean retention of at least 90\%. SQA denotes ScienceQA-IMG, and LIBERO denotes reconstructed LIBERO-10 offline QA.}
\label{tab:exp_main}
\begin{tabular*}{\linewidth}{@{\extracolsep{\fill}}lrrrrrrrrr@{}}
\toprule
 & & & \multicolumn{6}{c}{\textbf{Accuracy (\%)}} & \\
\cmidrule(lr){4-9}
\textbf{Method} & \shortstack{\textbf{Pruned}\\\textbf{layers}} & \shortstack{\textbf{Pruning}\\\textbf{(\%)}} & \textbf{AI2D} & \textbf{SQA} & \textbf{POPE} & \textbf{BLINK} & \textbf{ERQA} & \textbf{LIBERO} & \shortstack{\textbf{Mean ret.}\\\textbf{(\%)}} \\
\midrule
\rowcolor[gray]{0.94}
\multicolumn{10}{c}{\textbf{0.8B (24 layers)}} \\
\addlinespace[2pt]
OmniJev & - & 0.00 & 62.27 & 69.01 & 84.76 & 43.98 & 28.75 & 55.45 & 100.00 \\
\textbf{FastJEV} & 6 & 25.00 & 62.37 & 69.31 & 87.79 & 43.40 & 29.50 & 55.65 & 100.97 \\
\textbf{FastJEV} & 8 & 33.33 & 62.37 & 68.86 & 88.17 & 42.92 & 31.00 & 56.38 & 101.85 \\
\textbf{FastJEV} & \textbf{11} & \textbf{45.83} & 58.19 & 66.63 & 88.50 & 41.45 & 28.00 & 54.92 & \textbf{97.52} \\
\textbf{FastJEV} & 14 & 58.33 & 31.31 & 41.30 & 50.00 & 38.77 & 31.75 & 49.53 & 76.17 \\
\midrule
\rowcolor[gray]{0.94}
\multicolumn{10}{c}{\textbf{2B (24 layers)}} \\
\addlinespace[2pt]
OmniJev & - & 0.00 & 75.00 & 83.79 & 88.83 & 52.24 & 36.75 & 58.24 & 100.00 \\
\textbf{FastJEV} & 6 & 25.00 & 74.90 & 83.84 & 88.80 & 51.13 & 37.00 & 55.59 & 98.98 \\
\textbf{FastJEV} & 8 & 33.33 & 74.84 & 83.94 & 89.40 & 51.50 & 36.50 & 56.78 & 99.28 \\
\textbf{FastJEV} & \textbf{11} & \textbf{45.83} & 70.24 & 77.84 & 89.33 & 49.40 & 35.00 & 54.12 & \textbf{94.92} \\
\textbf{FastJEV} & 14 & 58.33 & 50.52 & 52.55 & 75.86 & 41.50 & 32.00 & 35.57 & 73.81 \\
\midrule
\rowcolor[gray]{0.94}
\multicolumn{10}{c}{\textbf{4B (32 layers)}} \\
\addlinespace[2pt]
OmniJev & - & 0.00 & 84.33 & 94.30 & 90.77 & 64.33 & 44.75 & 65.23 & 100.00 \\
\textbf{FastJEV} & 6 & 18.75 & 84.29 & 94.10 & 90.64 & 64.33 & 46.50 & 63.76 & 100.21 \\
\textbf{FastJEV} & 10 & 31.25 & 83.78 & 93.85 & 90.53 & 64.07 & 42.25 & 62.83 & 98.16 \\
\textbf{FastJEV} & \textbf{14} & \textbf{43.75} & 81.96 & 92.56 & 90.46 & 64.70 & 40.00 & 50.20 & \textbf{93.66} \\
\textbf{FastJEV} & 18 & 56.25 & 68.30 & 76.10 & 90.00 & 59.50 & 33.00 & 29.65 & 78.76 \\
\bottomrule
\end{tabular*}
\par\vspace{3pt}
\begin{minipage}{\linewidth}
\end{minipage}
\end{table*}

\paragraph{Quality under candidate depth reduction.}
~\cref{tab:exp_main} evaluates full FastJEV at four pruning budgets for each model scale.
Pruning 45.83\% of candidate layers gives mean task score retention of 97.52\% for 0.8B and 94.92\% for 2B.
For 4B, pruning 43.75\% retains 93.66\% of the Original score on average.
These results show that substantial candidate depth reduction is possible within the shared structure.
Increasing pruning beyond 56\% lowers mean retention to between 73.81\% and 78.76\% across the tested configurations.
The six-block rows use the default-budget experiment, while the larger budgets use their matched sweep references.

\paragraph{Quality at the default budget.}
At six skipped blocks, mean retention is 100.97\%, 98.98\%, and 100.21\% for 0.8B, 2B, and 4B, respectively.
The absolute accuracy change is below one percentage point in 13 of the 18 comparisons.
For 4B, AI2D changes from 84.33\% to 84.29\%, BLINK remains at 64.33\%, and ERQA improves from 44.75\% to 46.50\%.
LIBERO accuracy decreases by 2.66 and 1.46 percentage points for 2B and 4B.
Appendix~\ref{sec:exp_quality_details} reports episode bootstrap intervals and POPE F1 scores. \cref{sec:exp_pruning} examines task sensitivity at larger budgets.

\paragraph{From compactness to execution cost.}
At the default six-block budget, full FastJEV lowers total peak GPU memory in six of the 18 comparisons.
It is faster than Original on POPE at all three model scales and slower on the other five datasets.
On LIBERO, peak memory decreases by 15.70\%, 8.43\%, and 9.12\% for 0.8B, 2B, and 4B, respectively, with higher latency at each scale.
These measurements include the effects of state management, segmented execution, and readout scheduling.
At the larger budgets selected in \cref{tab:exp_main}, the geometric mean speedups over the six sets are 0.938, 0.939, and 0.953 for the three model scales.
\cref{tab:exp_costs} reports all default-budget measurements.
The following ablations examine how structural compactness relates to these execution costs.

\subsection{Component Interaction and Efficient Execution}
\label{sec:exp_components}

\begin{table*}[t]
\centering
\small
\setlength{\tabcolsep}{3pt}
\renewcommand{\arraystretch}{1.10}
\caption{Factorial comparison with 4B using a common recurrent kernel. C copies initial states, retains final caches, and executes full candidate depth. P skips six candidate blocks when enabled. C differs from the released Original in ~\cref{tab:exp_main}. LIBERO latency is per frame. Memory denotes the total allocated peak. Bold marks the best value in each column.}
\label{tab:exp_factors}
\begin{tabular*}{\linewidth}{@{\extracolsep{\fill}}lrrrrrr@{}}
\toprule
 & \multicolumn{3}{c}{\textbf{AI2D}} & \multicolumn{3}{c}{\textbf{LIBERO-10 QA}} \\
\cmidrule(lr){2-4}\cmidrule(lr){5-7}
\textbf{Mode} & Acc. (\%) $\uparrow$ & ms $\downarrow$ & MiB $\downarrow$ & Acc. (\%) $\uparrow$ & ms $\downarrow$ & MiB $\downarrow$ \\
\midrule
C & 84.23 & 132.51 & 9454.1 & \textbf{64.96} & 324.58 & 10381.8 \\
\midrule
T & 84.20 & 175.74 & 9525.0 & 64.89 & 838.51 & 9689.8 \\
S & 84.23 & 130.82 & 9261.6 & \textbf{64.96} & 319.20 & 9508.0 \\
P & \textbf{84.36} & 125.42 & 9453.0 & 63.90 & 296.14 & 10356.8 \\
\midrule
T+S & 84.20 & 174.98 & 9419.0 & 64.89 & 833.27 & 9477.9 \\
T+P & 84.29 & 163.45 & 9524.0 & 63.76 & 725.58 & 9685.9 \\
S+P & \textbf{84.36} & \textbf{125.05} & \textbf{9260.5} & 63.90 & \textbf{294.16} & 9483.9 \\
\midrule
\textbf{T+S+P} & 84.29 & 161.41 & 9419.2 & 63.76 & 721.34 & \textbf{9461.3} \\
\bottomrule
\end{tabular*}
\end{table*}

\paragraph{Interactions within the complete structure.}
We compare all eight combinations of T, S, and P with the same recurrent kernel.
The baseline C evaluates every candidate layer, copies the initial recurrent states, and retains the final caches.
It differs from the released Original.
~\cref{tab:exp_factors} presents the 4B results on AI2D and LIBERO, with P fixed to six skipped blocks when enabled.
With T and P fixed, S lowers total peak memory by 104.8 MiB on AI2D and 224.6 MiB on LIBERO, with unchanged accuracy.
With T and S fixed, P reduces latency on both datasets, while LIBERO accuracy changes from 64.89\% to 63.76\%.
S and P thus reduce state storage and executed depth within the shared candidate paths.
The comparison with and without T also changes segmentation and vocabulary projection scheduling, as detailed in Appendix~\ref{sec:exp_cost_details}.

\begin{figure*}[t]
\centering
\includegraphics[width=\textwidth]{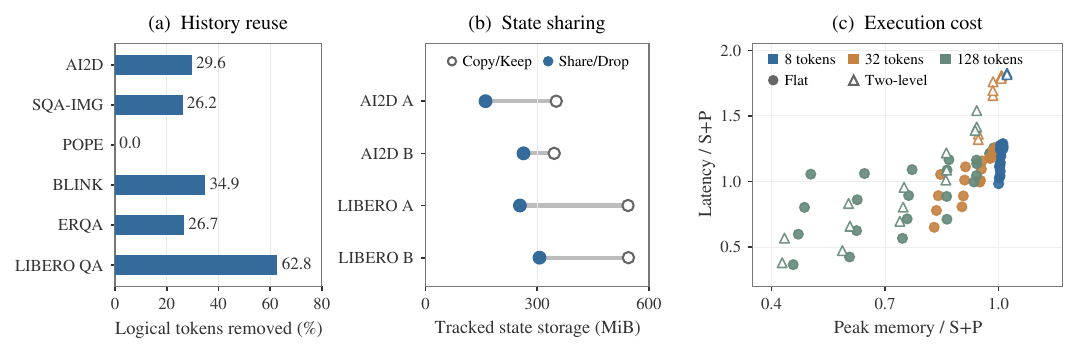}
\caption{Structural compactness and execution cost with 4B. (a) Logical candidate token reduction across the six complete evaluation sets. SQA-IMG denotes ScienceQA-IMG, and LIBERO QA is our reconstructed offline set. (b) Observed live state storage for four inputs, with T and six-layer P fixed. Copy / Keep and Share / Drop follow ~\cref{tab:exp_states}. (c) Full-request latency and total peak GPU memory of T+S+P relative to S+P across all 96 artificial workload conditions. Colors indicate candidate length, and marker shapes indicate branching structure. Ratios below one indicate lower cost. Artificial workloads evaluate execution cost only.}
\label{fig:exp_structure}
\end{figure*}

\paragraph{Understanding compactness and efficiency.}
T merges repeated candidate histories, S shares the states from which child paths continue, and P shortens the computation along these paths.
The complete structure combines these operations to reduce repeated histories, initial-state copies, and executed candidate layers.
~\cref{fig:exp_structure} relates history and state compactness to execution cost.
S+P removes history merging and evaluates candidate suffixes directly in batches.
Across six datasets and three model scales, it achieves $1.053\times$ to $1.258\times$ speedups over Original in all 18 comparisons.
Its 2B LIBERO accuracy changes from 58.24\% to 55.85\%.
Three independent processes on 4B AI2D and LIBERO preserve the same latency ordering, with S+P fastest, followed by Original and T+S+P.
Appendix~\ref{sec:exp_repeat_timing} reports the means and standard deviations.
We next examine the storage savings from S and the conditions under which the shared execution of T reduces cost.

\subsection{State Storage within the Shared Structure}
\label{sec:exp_states}

\begin{table*}[t]
\centering
\begin{minipage}[t]{0.42\linewidth}
\vspace{0pt}
\caption{Observed peak live state and cache storage (MiB) with 4B. A and B denote two representative inputs per dataset. T and six-layer P are fixed. Bold marks the lowest storage per input.}
\label{tab:exp_states}
\end{minipage}%
\hfill%
\begin{minipage}[t]{0.54\linewidth}
\vspace{0pt}
\caption{Layer selection accuracy (\%) with 4B and six skipped blocks in T+S+P. Random reports the mean and sample standard deviation over seeds 17, 29, and 43. Bold marks the best accuracy.}
\label{tab:exp_selection}
\end{minipage}
\par\vspace{2pt}
\begin{minipage}[t]{0.42\linewidth}
\vspace{0pt}
\centering
\small
\setlength{\tabcolsep}{2pt}
\renewcommand{\arraystretch}{1.10}
\begin{tabular*}{\linewidth}{@{\extracolsep{\fill}}llrrrr@{}}
\toprule
\multicolumn{2}{c}{\textbf{State policy}} & \multicolumn{2}{c}{\textbf{AI2D}} & \multicolumn{2}{c}{\textbf{LIBERO-10 QA}} \\
\cmidrule(lr){1-2}\cmidrule(lr){3-4}\cmidrule(lr){5-6}
Initial & Leaf & A & B & A & B \\
\midrule
Copy & Keep & 351.2 & 345.7 & 544.1 & 545.1 \\
Copy & Drop & 343.3 & 341.7 & 530.2 & 531.2 \\
\midrule
Share & Keep & 303.3 & 321.7 & 460.2 & 461.2 \\
Share & Drop & \textbf{161.3} & \textbf{263.2} & \textbf{253.7} & \textbf{306.2} \\
\bottomrule
\end{tabular*}
\end{minipage}%
\hfill%
\begin{minipage}[t]{0.54\linewidth}
\vspace{0pt}
\centering
\small
\setlength{\tabcolsep}{2pt}
\renewcommand{\arraystretch}{1.10}
\begin{tabular*}{\linewidth}{@{\extracolsep{\fill}}lrr@{}}
\toprule
 & \multicolumn{2}{c}{\textbf{Accuracy (\%) $\uparrow$}} \\
\cmidrule(lr){2-3}
Selection & AI2D & BLINK \\
\midrule
Random & $79.59 \pm 2.14$ & $61.56 \pm 2.93$ \\
Tail & 83.94 & 63.97 \\
Block Influence & 79.89 & \textbf{64.97} \\
Greedy Teacher KL & \textbf{84.42} & 64.12 \\
\midrule
\textbf{Output sensitivity} & 84.29 & 64.33 \\
\bottomrule
\end{tabular*}
\end{minipage}
\end{table*}

We fix T and the six-layer pruning mask and vary two state policies.
Initial recurrent states are copied or shared, and final recurrent caches at leaves are retained or omitted.
All four configurations retain internal trie states for continuation.
We use the AI2D inputs with the minimum and maximum segment counts in the timing subset and the first two frames in the LIBERO timing list.

\cref{tab:exp_states} counts live states and caches at program observation points, deduplicates their underlying storage, and reports the maximum observed value.
Share/Drop lowers this peak in all four cases.
For LIBERO A, it falls from 544.1 MiB with Copy/Keep to 253.7 MiB.
All 16 runs produce the same candidate choices and probabilities as the complete method on their corresponding input.

Accumulated over the LIBERO A request, initial-state copy allocations fall from 1,584 MiB to zero, and kernel final-state allocations fall from 1,254 to 380 MiB.
The remaining final states support the continuation of internal segments.
Initial-state sharing removes duplicate copies, while omitting leaf caches further reduces final-state allocation.
These measurements isolate the storage effect of S with the candidate histories and pruning mask held fixed.
The adjacent \cref{tab:exp_selection} compares layer selection criteria, discussed in~\cref{sec:exp_pruning}.

\subsection{History Reuse and Execution Structure}
\label{sec:exp_history}

\paragraph{Redundancy in real candidate sets.}
We count logical candidate token occurrences before padding with the 4B model.
T merges nodes only when their complete histories, positions, and execution conditions agree.
Across the complete evaluation sets, token counts fall by 29.64\% on AI2D, 26.16\% on ScienceQA-IMG, 34.88\% on BLINK, 26.74\% on ERQA, and 62.84\% on LIBERO.
POPE has no mergeable candidate histories under this protocol.

History merging also divides execution into segments.
LIBERO requires an average of 18.73 candidate segment calls per frame.
Profiling three frames gives 17 to 19 candidate backbone calls for T+S+P, compared with two for S+P.
On the timing subset, their latencies are 721.34 and 294.16 ms per frame, respectively.
The reduction in logical histories therefore coexists with additional calls between segments.

\begin{table*}[t]
\centering
\small
\setlength{\tabcolsep}{3pt}
\renewcommand{\arraystretch}{1.10}
\caption{Selected artificial 4B workloads with fixed source images and questions. Sharing is the fraction of candidate content held in common, excluding mandatory boundary tokens. Each condition uses four source inputs and five timing rounds. Speedup is S+P latency divided by T+S+P latency. Bold numbers mark the lower latency or peak memory in each pair. Artificial candidates are used to analyze execution cost.}
\label{tab:exp_workloads}
\begin{tabular*}{\linewidth}{@{\extracolsep{\fill}}rrrrrrrr@{}}
\toprule
\multicolumn{3}{c}{\textbf{Workload}} & \multicolumn{2}{c}{\textbf{Latency (ms) $\downarrow$}} &  & \multicolumn{2}{c}{\textbf{Peak memory (MiB) $\downarrow$}} \\
\cmidrule(lr){1-3}\cmidrule(lr){4-5}\cmidrule(lr){7-8}
Candidates & Tokens & Shared (\%) & S+P & T+S+P & Speedup $\uparrow$ & S+P & T+S+P \\
\midrule
2 & 8 & 0 & \textbf{110.54} & 140.13 & 0.789 & \textbf{9436.6} & 9549.0 \\
8 & 32 & 50 & \textbf{129.95} & 153.10 & 0.849 & 9646.7 & \textbf{9449.2} \\
\midrule
64 & 128 & 0 & \textbf{1511.67} & 1595.58 & 0.947 & 24742.3 & \textbf{12466.2} \\
64 & 128 & 75 & 1511.75 & \textbf{552.62} & 2.736 & 24615.1 & \textbf{11260.5} \\
\bottomrule
\end{tabular*}
\end{table*}

\paragraph{Candidate scale and shared content.}
We vary candidate counts, suffix lengths, shared content, and branching structure while fixing source images and questions.
The scan covers 96 conditions for 4B and 17 representative conditions for each smaller model.
These artificial candidates measure execution cost and are excluded from accuracy evaluation.
With S and P fixed, the trie implementation lowers latency in 27 of the 96 conditions for 4B and lowers total peak memory in 63.
It lowers latency in five conditions for each smaller model. \cref{tab:exp_workloads} shows four representative workloads.
With 64 candidates of 128 tokens and 75\% shared content, logical token occurrences fall from 8,192 to 2,207.
T+S+P achieves a $2.736\times$ speedup over S+P and lowers total peak memory by 54.25\%.
The complete structure can thus improve both costs when candidate evaluation contains sufficient shared computation.

These costs also reflect the readout implementation.
Trie readout projects at most 128 required nodes to the vocabulary at a time, while S+P projects all predictor positions in a candidate batch together.
With the same 64 candidates of 128 tokens but no shared content, total peak memory still falls by 49.62\%, while only 1.54\% of logical token occurrences are merged through common boundary markers.
The memory reduction measures the combined effect of trie execution and its readout schedule, rather than history merging alone.

\paragraph{The role of branching.}
We compare 24 pairs of flat and two-level tries with matched candidate counts, suffix lengths, and unique token counts.
The two-level trie is slower in 22 pairs, with latency ratios relative to the flat trie ranging from 0.950 to 1.499.
For eight candidates of 32 tokens, both structures contain 144 unique tokens.
Increasing the segment count from two to four changes latency from 153.10 to 218.79 ms.
The efficiency of shared execution thus depends on both the amount of reused history and its segmentation.

\subsection{Reducing the Depth of Shared Paths}
\label{sec:exp_pruning}

\begin{figure*}[t]
\centering
\begin{minipage}[t]{0.55\linewidth}
\vspace{0pt}
\centering
\includegraphics[width=0.95\linewidth]{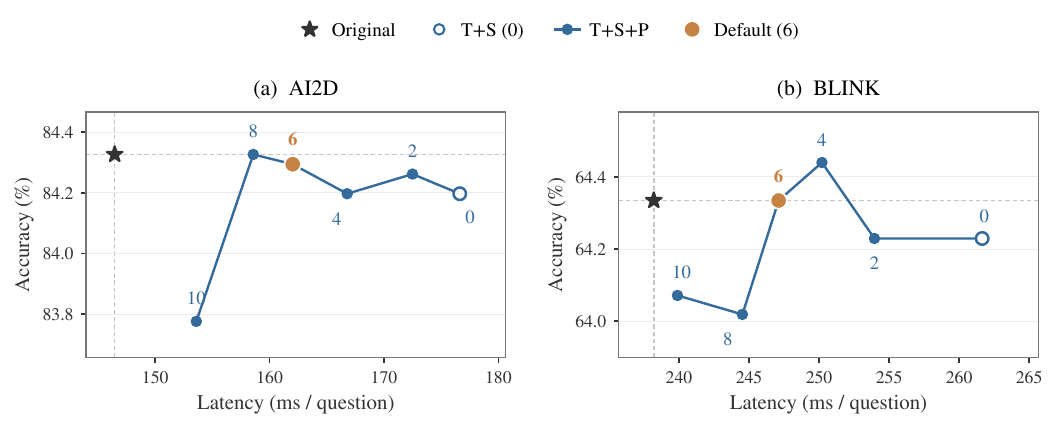}
\caption{Accuracy and latency of OmniJev 4B on AI2D and BLINK under fine-grained candidate pruning. Labels denote 0, 2, 4, 6, 8, or 10 skipped blocks, with T and S enabled. Stars mark Original measured in this sweep, open circles mark T+S without pruning, and orange points mark the six-block configuration. All points use complete evaluation splits.}
\label{fig:exp_pruning}
\end{minipage}%
\hfill%
\begin{minipage}[t]{0.43\linewidth}
\vspace{0pt}
\centering
\captionof{table}{Quality and execution cost of the 4B depth sweep over six datasets. Retention averages six accuracy ratios to the matched Original. Speedup divides Original latency by variant latency. Peak ratio divides variant total peak memory by Original peak memory. Both cost ratios use geometric means over the six datasets.}
\label{tab:exp_frontier}
\vspace{3pt}
\small
\setlength{\tabcolsep}{2pt}
\renewcommand{\arraystretch}{1.10}
\begin{tabular*}{\linewidth}{@{\extracolsep{\fill}}lrrrr@{}}
\toprule
Mode & Skip & Ret.(\%) & Speedup$\uparrow$ &Peak ratio $\downarrow$\\
\midrule
T+S+P & 10 & 98.16 & 0.890 & 0.994 \\
T+S+P & 14 & 93.66 & 0.953 & 0.993 \\
T+S+P & 18 & 78.76 & 1.011 & 0.993 \\
\midrule
S+P & 10 & 98.30 & 1.188 & 0.971 \\
S+P & 14 & 93.87 & 1.236 & 0.971 \\
S+P & 18 & 79.44 & 1.280 & 0.970 \\
\bottomrule
\end{tabular*}
\end{minipage}
\end{figure*}

\paragraph{Fine-grained pruning behavior.}
~\cref{fig:exp_pruning} plots accuracy against latency on AI2D and BLINK with 4B.
We retain T and S and vary the number of skipped candidate blocks from zero to ten in steps of two.
Increasing pruning lowers latency on both tasks, while accuracy changes non-monotonically.
This local variation motivates examining larger budgets across all six datasets.

\paragraph{Exploring the pruning range.}
We use the fixed sensitivity rankings to examine how much candidate depth can be removed.
For 4B, we first screen nine budgets on 128 fixed questions per public dataset and all 188 LIBERO frames.
For T+S+P and S+P separately, we select the largest tested budget with at least 90\% mean retention and its two neighboring budgets.
Their union gives 10, 14, and 18 skipped blocks, which we evaluate on all six complete splits.
For 0.8B and 2B, the complete-split sweep tests 6, 8, 11, and 14 skipped blocks.
The six-block rows in \cref{tab:exp_main} retain the default-budget results, and the larger-budget rows use the sweep results.
This analysis uses the evaluation sets to characterize compression limits.
The 90\% criterion selects budgets from measured task scores, while the layer rankings remain fixed without target labels.

\paragraph{Task sensitivity under pruning.}
The larger-budget results in \cref{tab:exp_main} show that tasks respond differently to pruning.
At ten skipped blocks, full FastJEV retains at least 94.41\% on every dataset with 4B.
At fourteen blocks, POPE and BLINK retain 99.66\% and 100.57\%, while ERQA and LIBERO retain 89.39\% and 76.96\%.
Four datasets reach 90\% retention with T+S+P, and five with S+P.
For 0.8B and 2B at eleven skipped blocks, all six datasets exceed 90\%, with minima of 93.45\% and 92.61\%.
These differences show that the pruning range depends on both task and model scale.

\paragraph{Execution cost under deeper pruning.}
\cref{tab:exp_frontier} compares quality and cost for both 4B configurations.
Latency and memory are measured against Original within the same process using the protocol in \cref{sec:exp_setup}.
Cost ratios are normalized within each dataset and aggregated by their geometric mean.
At fourteen skipped blocks, mean retention is 93.66\% for T+S+P and 93.87\% for S+P.
Their speedups over Original are $0.953\times$ and $1.236\times$, with total peak memory ratios of 0.993 and 0.971.
Fourteen blocks is the largest fully evaluated budget meeting the mean retention criterion for both configurations.
For 0.8B and 2B at eleven skipped blocks, speedups are $0.938\times$ and $0.939\times$, with peak memory ratios of 0.985 and 1.023.
Pruning reduces computation along candidate paths, while context encoding and the execution of these paths continue to contribute to request latency.

\paragraph{Selection at a fixed budget.}
We compare output sensitivity with Block Influence based on hidden-state similarity~\citep{shortgpt2025}, tail pruning, three random masks, and greedy teacher KL selection.
All methods use the same six-block budget, calibration source, and T+S+P executor.
Random masks use seeds 17, 29, and 43, and report mean accuracy and sample standard deviation.
\cref{tab:exp_selection} reports complete-split accuracy on AI2D and BLINK.
Output sensitivity reaches 84.29\% on AI2D, close to greedy teacher KL at 84.42\% and above Block Influence, tail pruning, and the random mean.
On BLINK, it reaches 64.33\%, below Block Influence at 64.97\% and above the other criteria.
The two tasks favor different criteria, while output sensitivity remains competitive on both.
It provides the fixed ranking reused across tasks and pruning budgets in our experiments.

\subsection{Discussion}
\label{sec:exp_discussion}

\paragraph{Does more sharing always mean faster inference?}
Prefix reuse is a natural design for JEV inference, where candidates are evaluated against a common context. Existing context reuse avoids repeatedly encoding this shared input~\citep{omnijev2026}. Extending reuse to matching candidate histories further reduces repeated computation. However, our OmniJev experiments show that this extension can increase inference latency. Independent batched candidate evaluation can execute faster despite retaining repeated histories. Shared context anchoring with decision guided pruning (S+P) retains common context reuse. The comparison therefore examines the benefit of additional sharing within candidate suffixes. The finding highlights a gap between reducing repeated work and reducing execution time.

\paragraph{Why can less computation take more time?}
In our trie executor, candidate paths are divided into segments at forks, and continuation states connect these segments. Frequent branching can leave little computation within each segment, while traversal, kernel launches, and state handling still incur costs. Independent candidate evaluation keeps complete suffixes in regular batches. Sharing therefore changes both the amount of computation and the opportunities to batch it. Its execution benefit depends on whether the saved computation outweighs the cost of segmentation and state handling. Pruning further reduces computation within each segment without removing its boundaries, which may make these costs more prominent.

\paragraph{What does this imply for JEV design?}
Full FastJEV connects history reuse, state storage, and candidate depth in one compact structure. Its comparison with this ablation shows why the extent of sharing alone is insufficient to assess execution efficiency. The useful granularity of sharing also depends on branching structure and the computation remaining after pruning. This motivates future schedules that preserve shared histories while grouping compatible continuations into larger batches. Our findings suggest evaluating prefix sharing together with its execution schedule when designing efficient JEV models.

\section{Conclusion}
\label{sec:conclusion}

In this paper, we presented FastJEV for compact candidate evaluation in JEV models. Shared context anchoring and candidate prefix sharing jointly organize state storage and matching token histories. Decision guided pruning then reduces candidate depth according to its effect on final scores. This structure retains full context encoding and all candidates without additional training. Across three OmniJev v1.1 scales, the complete method removes about 44\% to 46\% of candidate layers. These configurations retain over 93\% of the original task scores on average across six evaluation sets. Controlled experiments show that the complete structure reduces repeated histories and state storage. Combining shared context anchoring with decision guided pruning improves efficiency when candidate suffixes are evaluated independently in batches. The execution benefit of history sharing depends on candidate overlap and branching structure. These findings provide a structural view of JEV inference and a basis for further study of compact candidate evaluation.

\bibliographystyle{plainnat}
\bibliography{references}

\newpage
\appendix

\section*{Supplementary Material}

This supplement provides complete experimental results and extended pruning analyses for FastJEV.
All experiments use OmniJev v1.1.
LIBERO results refer to the reconstructed offline question set described in the main paper.

\section{Additional Experimental Results}
\label{sec:exp_cost_details}

\cref{tab:exp_costs} reports the complete quality, latency, and peak GPU memory measurements for the six-block main evaluation.
It retains all three model scales and six datasets from \cref{tab:exp_main}.
FastJEV uses T+S+P in every comparison.
The measurements follow \cref{sec:exp_setup}, including full model weights in peak memory.
The larger-budget sweep uses its own matched Original measurements and is reported separately in \cref{tab:exp_frontier}.

Trie readout projects at most 128 required nodes to the vocabulary at a time, while the independent branch path projects all predictor positions in a candidate batch together.
The measured cost differences between T+S+P and S+P therefore include history merging, segment scheduling, and vocabulary projection scheduling.
Their individual cost contributions are not isolated by these comparisons.

The complete method lowers total peak memory in six of the 18 comparisons.
At all three model scales, it is faster than Original on POPE and slower on the other five datasets.
POPE has no candidate histories that can be merged under this protocol, so its speedup reflects other differences in candidate execution.
The LIBERO memory reductions of 15.70\%, 8.43\%, and 9.12\% occur with higher latency.
\cref{sec:exp_history} examines segmentation and readout scheduling alongside logical history reduction.

\begin{table*}[t]
\centering
\small
\setlength{\tabcolsep}{3pt}
\renewcommand{\arraystretch}{1.08}
\caption{Complete accuracy and inference cost at the default six-block budget across three OmniJev model sizes. FastJEV uses all three components (T+S+P). Bold marks the best value in each pair at the displayed precision. Latency is per question, except for LIBERO-10 QA, which is per frame of eight questions. Peak GPU memory includes model weights and uses the maximum PyTorch allocated peak over the measured inputs. LIBERO-10 QA is our reconstructed offline set.}
\label{tab:exp_costs}
\begin{tabular*}{\linewidth}{@{\extracolsep{\fill}}lrrrrrr@{}}
\toprule
 & \multicolumn{2}{c}{\textbf{Accuracy (\%) $\uparrow$}} & \multicolumn{2}{c}{\textbf{Latency (ms) $\downarrow$}} & \multicolumn{2}{c}{\textbf{Peak memory (MiB) $\downarrow$}} \\
\cmidrule(lr){2-3}\cmidrule(lr){4-5}\cmidrule(lr){6-7}
\textbf{Dataset} & Original & \textbf{FastJEV} & Original & \textbf{FastJEV} & Original & \textbf{FastJEV} \\
\midrule
\multicolumn{7}{@{}l}{\textbf{0.8B}} \\
\addlinespace[1pt]
AI2D & 62.27 & \textbf{62.37} & \textbf{98.09} & 105.54 & \textbf{1930.9} & 1934.8 \\
ScienceQA-IMG & 69.01 & \textbf{69.31} & \textbf{100.17} & 106.60 & 1907.9 & \textbf{1890.8} \\
POPE & 84.76 & \textbf{87.79} & 91.26 & \textbf{75.30} & \textbf{1832.1} & 1877.8 \\
BLINK & \textbf{43.98} & 43.40 & \textbf{119.35} & 123.10 & \textbf{2020.2} & 2075.1 \\
ERQA & 28.75 & \textbf{29.50} & \textbf{111.84} & 133.41 & \textbf{2133.5} & 2219.4 \\
LIBERO-10 QA & 55.45 & \textbf{55.65} & \textbf{192.44} & 487.10 & 2312.8 & \textbf{1949.8} \\
\addlinespace[3pt]
\midrule
\multicolumn{7}{@{}l}{\textbf{2B}} \\
\addlinespace[1pt]
AI2D & \textbf{75.00} & 74.90 & \textbf{111.99} & 120.30 & \textbf{4549.1} & 4644.0 \\
ScienceQA-IMG & 83.79 & \textbf{83.84} & \textbf{111.14} & 121.44 & \textbf{4523.6} & 4603.7 \\
POPE & \textbf{88.83} & 88.80 & 104.29 & \textbf{89.27} & \textbf{4464.8} & 4590.4 \\
BLINK & \textbf{52.24} & 51.13 & \textbf{163.29} & 167.12 & \textbf{4633.2} & 4930.3 \\
ERQA & 36.75 & \textbf{37.00} & \textbf{141.95} & 163.36 & \textbf{4753.9} & 5220.9 \\
LIBERO-10 QA & \textbf{58.24} & 55.59 & \textbf{235.84} & 538.90 & 5012.7 & \textbf{4590.4} \\
\addlinespace[3pt]
\midrule
\multicolumn{7}{@{}l}{\textbf{4B}} \\
\addlinespace[1pt]
AI2D & \textbf{84.33} & 84.29 & \textbf{145.33} & 161.41 & 9454.0 & \textbf{9419.2} \\
ScienceQA-IMG & \textbf{94.30} & 94.10 & \textbf{137.39} & 156.90 & 9387.3 & \textbf{9340.2} \\
POPE & \textbf{90.77} & 90.64 & 126.82 & \textbf{105.92} & \textbf{9192.5} & 9322.5 \\
BLINK & \textbf{64.33} & \textbf{64.33} & \textbf{246.36} & 253.48 & \textbf{9629.2} & 9792.3 \\
ERQA & 44.75 & \textbf{46.50} & \textbf{202.21} & 238.19 & \textbf{9788.3} & 10163.1 \\
LIBERO-10 QA & \textbf{65.23} & 63.76 & \textbf{336.46} & 721.34 & 10410.2 & \textbf{9461.3} \\
\bottomrule
\end{tabular*}
\end{table*}

\subsection{Independent Timing Runs}
\label{sec:exp_repeat_timing}

We repeat the 4B measurements on AI2D and LIBERO in three independent processes with fixed inputs and configurations.
Each process uses five timing rounds on AI2D and three on LIBERO.
We report the mean and sample standard deviation of the three process means.
On AI2D, latency is $144.98\pm0.56$ ms for Original, $124.42\pm0.26$ ms for S+P, and $161.56\pm1.59$ ms for T+S+P.
On LIBERO, the corresponding values are $336.27\pm2.45$, $295.09\pm2.16$, and $723.27\pm12.29$ ms per frame.
All three processes preserve the same latency ordering on both datasets.

Profiling three 4B LIBERO frames gives 4,963 to 5,027 kernel launches within candidate computation for S+P and 41,647 to 47,271 for T+S+P.
These counts complement the backbone call analysis in \cref{sec:exp_history} and expose the finer execution granularity of the trie path.

\subsection{Additional Decision Quality Measures}
\label{sec:exp_quality_details}

At the default six-block budget, POPE F1 scores for Original and T+S+P are 86.05\% and 88.29\% for 0.8B, 89.00\% and 89.00\% for 2B, and 90.52\% and 90.50\% for 4B.
On reconstructed LIBERO-10 QA, the accuracy changes for 2B and 4B are $-2.66$ and $-1.46$ percentage points.
Using 2,000 paired bootstrap samples over the 20 evaluation episodes, their 95\% confidence intervals are $[-4.12,-1.27]$ and $[-2.76,-0.07]$ percentage points.
The resampling unit is an episode, preserving the grouping of its frames and questions.

\subsection{Understanding Compact Candidate Evaluation}
\label{sec:supp_understanding}

\paragraph{Shared histories and execution granularity.}
T, S, and P act on connected parts of candidate evaluation.
T organizes matching histories into shared paths, S reduces the state copies needed to execute their continuations, and P shortens the depth of these paths.
\cref{fig:exp_structure} of the main paper measures the resulting reductions in repeated histories and live state storage.
The branching experiment further shows that token reuse alone does not determine execution cost.
With matched unique token counts, the two-level trie is slower than the flat trie in 22 of 24 pairs.
Additional segments introduce more calls between shared paths and their continuations.
The efficiency of T therefore depends on both the amount of shared work and how that work is organized.
S+P retains state sharing and pruning while evaluating candidate suffixes directly in batches, providing a complementary view of their execution efficiency.

\paragraph{State storage and total memory.}
The state experiment fixes T and the pruning mask while changing state storage policies.
For LIBERO A, sharing initial states and omitting unused leaf caches lowers observed live state storage from 544.1 to 253.7 MiB without changing the candidate probabilities.
This isolates a storage benefit within the same candidate structure.
Total GPU peak memory also includes model weights, activations, attention caches, and readout buffers.
All model weights remain resident because context encoding retains the full backbone.
The reduction in candidate state storage therefore accounts for only part of the total memory footprint.
The readout scheduling difference described above also contributes to the cost comparison between trie and independent execution.

\paragraph{Takeaway.}
The complete structure jointly reduces repeated histories, state copies, and executed candidate layers.
Its latency benefit depends on the amount of shared work and the granularity of execution.

\subsection{Qualitative Comparisons}
\label{sec:supp_qualitative}

\begin{figure*}[p]
\centering
\includegraphics[width=\textwidth,height=0.80\textheight,keepaspectratio]{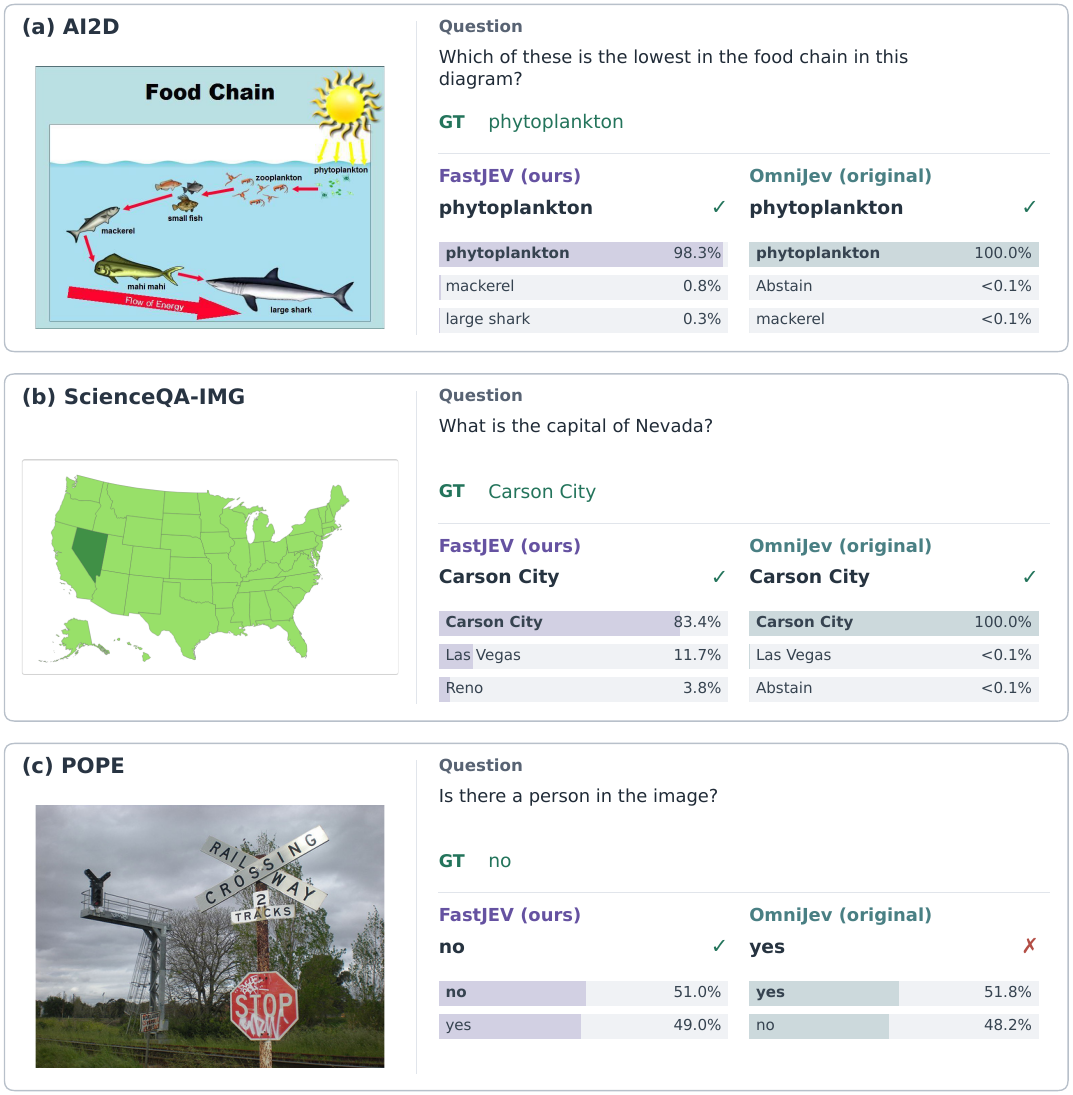}
\caption{Selected correct FastJEV answers on AI2D, ScienceQA-IMG, and POPE using OmniJev 4B. FastJEV uses shared context anchoring, candidate prefix sharing, and pruning of 14 candidate layers. Each panel places the image on the left and the question and answer comparison on the right. GT denotes the evaluation label. Bars show up to three highest probabilities from each model without renormalization. Checks and crosses indicate agreement with GT.}
\label{fig:exp_qualitative_general}
\end{figure*}

\begin{figure*}[p]
\centering
\includegraphics[width=\textwidth,height=0.80\textheight,keepaspectratio]{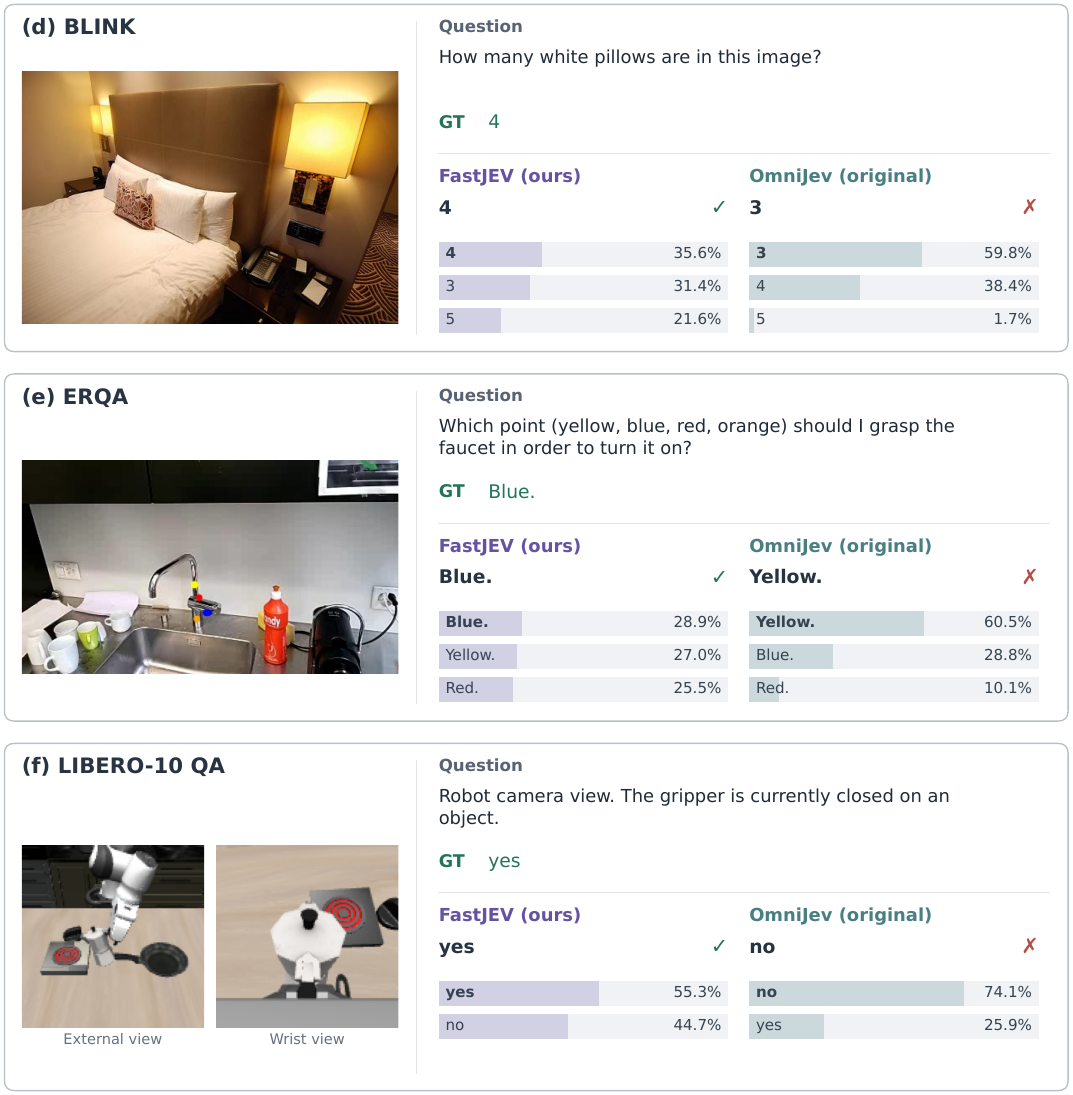}
\caption{Selected correct FastJEV answers on BLINK, ERQA, and reconstructed LIBERO-10 offline QA under the same OmniJev 4B configuration. All input views are shown, including both robot cameras. Answers display the selected candidate text. Appended choice lists and response-format instructions are omitted from displayed questions. The LIBERO ground truth uses the frozen state-derived label. Bars retain the original model probabilities and are not normalized over the displayed subset.}
\label{fig:exp_qualitative_spatial}
\end{figure*}

\cref{fig:exp_qualitative_general} and~\cref{fig:exp_qualitative_spatial} compare individual decisions from OmniJev and FastJEV across the six evaluation sets.
Both use the 4B v1.1 model.
FastJEV combines all three components and skips 14 of the 32 candidate layers, while retaining full context encoding.
We show the input images, question, ground truth, and selected answers alongside the highest candidate probabilities.
We select examples answered correctly by FastJEV, including cases where both models are correct and cases where their answers differ.
Aggregate performance is reported in \cref{tab:exp_main}.

\paragraph{Preserved decisions and changing scores.}
The diagram and science examples retain the correct answer after candidate depth is reduced by 43.75\%.
Their probability distributions change even when the selected answer remains the same.
The object presence and counting examples show changes from an incorrect to a correct answer.
In the counting example, the probability of the correct answer decreases slightly, but its rank rises because the competing answer decreases more.
This example illustrates the role of relative candidate scores in the final decision.

\paragraph{Correct decisions across visual tasks.}
The remaining examples extend this comparison to object interaction and robot state recognition.
FastJEV selects the marked faucet handle in ERQA and correctly identifies that the robot gripper holds an object in the reconstructed LIBERO question.
Together, these examples illustrate correct candidate decisions across diagrams, scene images, and robot observations with reduced candidate depth.

\section{Extended Pruning Analysis}
\label{sec:supp_pruning}

\subsection{Stability of Layer Selection}
\label{sec:supp_calibration_size}

Decision guided pruning estimates layer sensitivity from a small set of unlabeled inputs. We examine whether changing the size of this set changes the layers selected for pruning. We evaluate the 0.8B, 2B, and 4B OmniJev models using three disjoint calibration draws from OK-VQA. Each draw contains 128 inputs and defines nested subsets of 8, 16, 32, and 64 inputs. All three models use the same calibration draws. Each input contains 64 candidate answers, and no answer labels are used for layer selection. We fix the pruning budget at six candidate blocks and retain all blocks for context encoding. \cref{tab:supp_calibration_size} reports the selected blocks for each calibration set size. For each model, the selected set remains unchanged across all five sizes and all three draws. These sets also match the corresponding masks used in the main experiment.
Although the models select different blocks, each model recovers a consistent set from the tested calibration subsets.

\begin{table*}[t]
\centering
\small
\setlength{\tabcolsep}{10pt}
\renewcommand{\arraystretch}{1.15}
\caption{Selected candidate blocks across calibration set sizes for OmniJev v1.1. Each entry lists the six blocks selected for pruning. All three calibration draws yield the same set in every entry. Layer indices start at one and are listed in ascending order. Context encoding retains all blocks.}
\label{tab:supp_calibration_size}
\begin{tabular}{cccc}
\toprule
Calibration inputs & 0.8B & 2B & 4B \\
\midrule
8 & $\{16,17,18,19,20,21\}$ & $\{15,17,18,19,21,22\}$ & $\{21,23,24,25,26,27\}$ \\
16 & $\{16,17,18,19,20,21\}$ & $\{15,17,18,19,21,22\}$ & $\{21,23,24,25,26,27\}$ \\
32 & $\{16,17,18,19,20,21\}$ & $\{15,17,18,19,21,22\}$ & $\{21,23,24,25,26,27\}$ \\
64 & $\{16,17,18,19,20,21\}$ & $\{15,17,18,19,21,22\}$ & $\{21,23,24,25,26,27\}$ \\
128 & $\{16,17,18,19,20,21\}$ & $\{15,17,18,19,21,22\}$ & $\{21,23,24,25,26,27\}$ \\
\bottomrule
\end{tabular}
\end{table*}

\subsection{Pruning Budgets and Depth Redundancy}

This section supplements the pruning results in \cref{fig:exp_pruning} and \cref{tab:exp_frontier} of the main paper.
We describe budget selection and reference measurements, then discuss candidate depth redundancy.

\paragraph{Budget selection and reference measurements.}
The layer ranking is fixed from calibration before the budget scan.
For 4B, the initial screen evaluates nine budgets using 128 fixed questions per public dataset and all 188 LIBERO frames.
The largest tested budget meeting 90\% mean retention and its neighbors determine the budgets evaluated on complete splits.
This scan characterizes compression on the reported evaluation sets.
Both the fine-grained sweep and the larger-budget sweep use their own matched Original measurements.
The same verified Original accuracies normalize all points within the larger-budget sweep.
Latency and peak memory are measured against Original within the same process for each evaluated budget and dataset.

For 0.8B and 2B, the complete-split scan evaluates 6, 8, 11, and 14 skipped blocks.
At 11 blocks, all six datasets exceed 90\% retention, with minima of 93.45\% and 92.61\%, respectively.
\cref{tab:exp_main} of the main paper reports the default-budget results at six blocks and the sweep results at larger budgets.

\paragraph{Understanding depth redundancy.}
Context encoding remains at full depth throughout these sweeps.
The retained quality at roughly 44\% to 46\% candidate depth reduction therefore supports reducing candidate computation after the context has been fully encoded.
At a fixed pruning budget, T+S+P and S+P use the same layer mask and show similar retention curves.
For example, their mean retention differs by only 0.21 percentage points at 14 skipped blocks with 4B.
This similarity suggests that the task differences in this sweep are mainly associated with the candidate depth removed under both execution structures.
The larger retention losses on ERQA and LIBERO indicate a narrower pruning range for these tasks than for POPE and BLINK.

\paragraph{Takeaway.}
Candidate depth can be reduced substantially while preserving full context encoding.
The task curves identify where deeper pruning begins to reduce decision quality within the tested range.

\section{Limitations and Future Work}
\label{sec:supp_limitations}

FastJEV provides a structural perspective on redundancy in JEV candidate evaluation.
The current implementation introduces scheduling overhead when shared paths are split into short segments.
Our analysis is based on OmniJev v1.1.
Further work is needed to reduce this overhead and examine how the findings extend to more JEV models.

Our pruning strategy uses decision sensitivity to reduce candidate depth, while token pruning reduces sequence length.
These two dimensions offer complementary opportunities for compression.
Future work can explore their combination and evaluate its effects on decision quality and execution cost.


\end{document}